\documentclass[letterpaper]{article} 
\usepackage{aaai2026}  
\usepackage{times}  
\usepackage{helvet}  
\usepackage{courier}  
\usepackage[hyphens]{url}  
\usepackage{graphicx} 
\usepackage{natbib}  
\usepackage{caption} 
\usepackage{algorithm}
\usepackage{algorithmic}

\usepackage{booktabs} 
\usepackage[table]{xcolor}

\usepackage{newfloat}
\usepackage{listings}
\DeclareCaptionStyle{ruled}{labelfont=normalfont,labelsep=colon,strut=off} 
\floatstyle{ruled}
\newfloat{listing}{tb}{lst}{}
\floatname{listing}{Listing}
\title{Agentic DAG-Orchestrated Planner Framework for Multi-Modal, Multi-Hop Question Answering in Hybrid Data Lakes}
\author {
    Kirushikesh D B \textsuperscript{\rm 1},
    Manish Kesarwani\textsuperscript{\rm 1},
    Nishtha Madaan\textsuperscript{\rm 1},
    Sameep Mehta\textsuperscript{\rm 1},\\
    Aldrin Dennis\textsuperscript{\rm 2},
    Siddarth Ajay\textsuperscript{\rm 2},
    Rakesh B R\textsuperscript{\rm 2},    
    Renu Rajagopal\textsuperscript{\rm 2},
    Sudheesh Kairali\textsuperscript{\rm 2}
}
\affiliations {
    \textsuperscript{\rm 1}IBM Research, India, 
    \textsuperscript{\rm 2}IBM Software, India\\
    kirushi@ibm.com, 
    \{manishkesarwani, nishthamadaan, sameepmehta\}@in.ibm.com, \\
    \{Aldrin.Dennis1, siddarth.ajay, raki, Renu.Rajagopal\}@ibm.com, 
    skairali@in.ibm.com
}

\usepackage{bibentry}
\usepackage{amsmath}
\usepackage{amssymb}

\usepackage{makecell}

\begin{document}

\maketitle

\begin{abstract}
Enterprises increasingly need natural language (NL) question answering over hybrid data lakes that combine structured tables and unstructured documents. Current deployed solutions, including RAG-based systems, typically rely on brute-force retrieval from each store and post-hoc merging. Such approaches are inefficient and leaky, and more critically, they lack explicit support for multi-hop reasoning, where a query is decomposed into successive steps (hops) that may traverse back and forth between structured and unstructured sources.

We present {\bf A}gentic {\bf D}AG-{\bf O}rchestrated {\bf T}ransformer (A.DOT) Planner, a framework for multi-modal, multi-hop question answering, that compiles user NL queries into directed acyclic graph (DAG) execution plans spanning both structured and unstructured stores. The system decomposes queries into parallelizable sub-queries, incorporates schema-aware reasoning, and applies both structural and semantic validation before execution. The execution engine adheres to the generated DAG plan to coordinate concurrent retrieval across heterogeneous sources, route intermediate outputs to dependent sub-queries, and merge final results in strict accordance with the plan’s logical dependencies. Advanced caching mechanisms, incorporating paraphrase-aware template matching, enable the system to detect equivalent queries and reuse prior DAG execution plans for rapid re-execution, while the DataOps System addresses validation feedback or execution errors. The proposed framework not only improves accuracy and latency, but also produces explicit evidence trails, enabling verification of retrieved content, tracing of data lineage, and fostering user trust in the system’s outputs.

On benchmark dataset, A.DOT achieves 14.8\% absolute gain in correctness and 10.7\% in completeness over baselines. The system is currently under evaluation for deployment as the data retrieval module in IBM Watsonx.data Premium, with planned user studies, large-scale performance testing, and validation over representative client datasets, establishing a clear and immediate trajectory from research prototype to enterprise deployment.
\end{abstract}


\section{Introduction}
\label{sec:intro}

Hybrid data lakes, which co-locate structured tables with unstructured document repositories, are increasingly used in enterprise environments. By integrating structured enterprise records (e.g., invoice metadata) with unstructured documents (e.g., full invoice texts, contracts), organizations can support diverse analytical workloads and enable decision-making across different business units. Business users increasingly expect to query such heterogeneous sources in natural language, without requiring knowledge of SQL, vector-search syntax, or custom extraction pipelines. An effective system must interpret user intent, locate relevant data across modalities, and synthesize a coherent answer.

\begin{figure}[ht]
  \centering
  \includegraphics[width=0.45\textwidth]{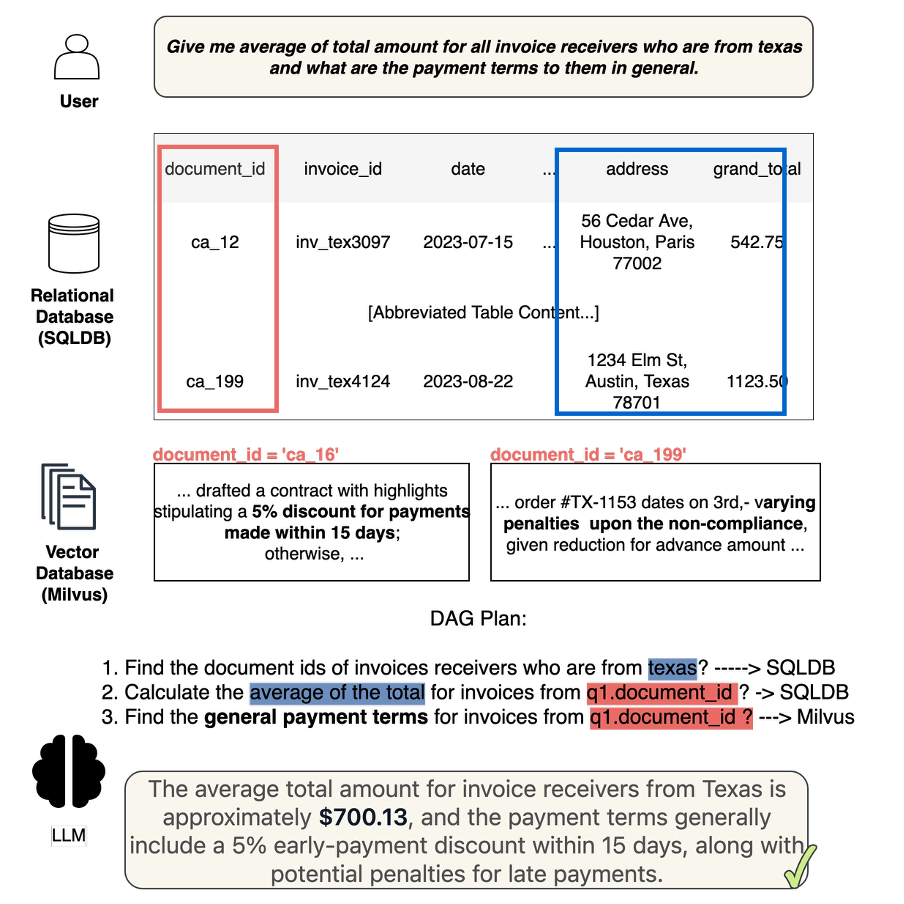}
  \caption{Question Answering over Hybrid Data Lakes}
  \label{fig:motivation_example}
\end{figure}

Figure~\ref{fig:motivation_example} shows an example enterprise query: \textit{``Give me the average total amount for invoice receivers from Texas and their general payment terms.''} Answering this requires (i) identifying invoices with Texas addresses in structured tables, (ii) computing the average grand total for those invoices, and (iii) retrieving payment terms from associated unstructured invoice documents stored in a vector database. 

State-of-the-art systems \cite{bardhan2024ttqa} would submit the NL query independently to both structured and unstructured stores. This design introduces several inefficiencies. First, it places heavy reliance on NL2SQL (or NL2Vector) models to generate precise queries, which increases the risk of hallucinations \cite{qu2024before}, especially when key information (e.g., payment terms) is not present in the underlying data source at all. Second, to avoid missing relevant results, these systems often over-retrieve, pulling back large volumes of data. This not only increases computational overhead but can also cause unnecessary data leakage, exposing more information than required.

In contrast, a more principled approach is to fetch only the precise data required from each source, while maintaining explicit data lineage and selectively passing intermediate results across modalities. Such targeted execution reduces unnecessary overhead, mitigates data leakage risks, and, crucially for enterprise settings, enables provenance tracking so that users can verify the origin of every returned answer.


Benchmarks such as HybridQA~\cite{hybridqa} and MMQA~\cite{mmqa2021} highlight the need for multi-hop reasoning across heterogeneous sources: HybridQA contains ~7K questions (in the dev and test set) requiring reasoning over tables and linked passages while MMQA extends the setting to text, tables, and images, showing the difficulty of joint reasoning beyond two modalities. Table-oriented models (e.g., TaBERT~\cite{yin2020tabert}, TAPAS~\cite{herzig2020tapas}) and schema-aware text-to-SQL systems such as RAT-SQL~\cite{wang2019rat} focus on single-modal pipelines and typically produce extractive answers. Agentic reasoning frameworks like ReAct~\cite{yao2023react} interleave reasoning with tool calls but execute sub-tasks sequentially, incurring higher latency. DAG-based planners such as LLM-Compiler~\cite{kim2024llm} compile dependency graphs but often replace structured intermediates with raw text, which is inefficient when downstream steps require only subsets of results. A comprehensive review of related work is presented in Appendix~\ref{app:rel_works}. To our knowledge, no existing approach supports parallel execution, plan caching, variable-scoped data passing, and evidence traceability in a unified framework; moreover, multi-hop reasoning that repeatedly alternates between structured and unstructured data remains unaddressed.

In this paper, we present the \textbf{A}gentic \textbf{D}AG-\textbf{O}rchestrated \textbf{T}ransformer (\textbf{A.DOT}) Planner, a framework for multi-modal, multi-hop question answering in hybrid data lakes. A.DOT compiles the input NL query into a DAG plan, where each node corresponds to an atomic sub-query mapped to the appropriate data source (e.g., relational database or vector store). A single LLM pass generates the plan, which is then validated both structurally and semantically. If inconsistencies or errors are detected, the DataOps System intervenes, first at the planning stage to refine the DAG against schema and semantic constraints, and later at execution time if the engine raises runtime errors. Independent nodes are executed in parallel, with intermediate variables passed selectively to dependent steps. Paraphrase-aware caching enables the reuse of prior DAG plans for semantically equivalent queries, reducing redundant computation. Finally, the plan-driven execution maintains explicit data provenance across all intermediate stages, producing evidence trails that support verification, lineage tracking, and compliance auditing.

A.DOT is being evaluated for deployment as the primary retrieval module in \textit{IBM Watsonx.data Premium} \cite{think2025}, with the goal of improving answer accuracy, and providing verifiable outputs in enterprise settings.

The contributions of this work are as follows:
\begin{enumerate}
    \item We propose A.DOT, an agentic framework for question answering over hybrid data lakes, combining DAG-based plan generation with structural and semantic validation, a dual-stage DataOps system for diagnosis and remediation, paraphrase-aware plan caching, parallel execution, and fine-grained lineage tracking.
    \item A variable-binding mechanism for inter-node communication that selectively passes only the required data elements between sub-queries across both structured and unstructured sources, thereby improving scalability and reducing data leakage over large enterprise data lakes.
    \item A parallel execution strategy that exploits DAG independence to reduce end-to-end latency, while enabling intermediate result exposure to enhance responsiveness.
    \item A plan-driven execution model that records evidence and data lineage, enabling verification and auditability.
    \item Empirical evaluation on HybridQA dataset demonstrates a 14.8\% improvement in answer correctness and a 10.7\% gain in completeness over strong agentic baselines.
\end{enumerate}



\begin{figure*}[ht!]
    \centering
    \includegraphics[width=\textwidth]{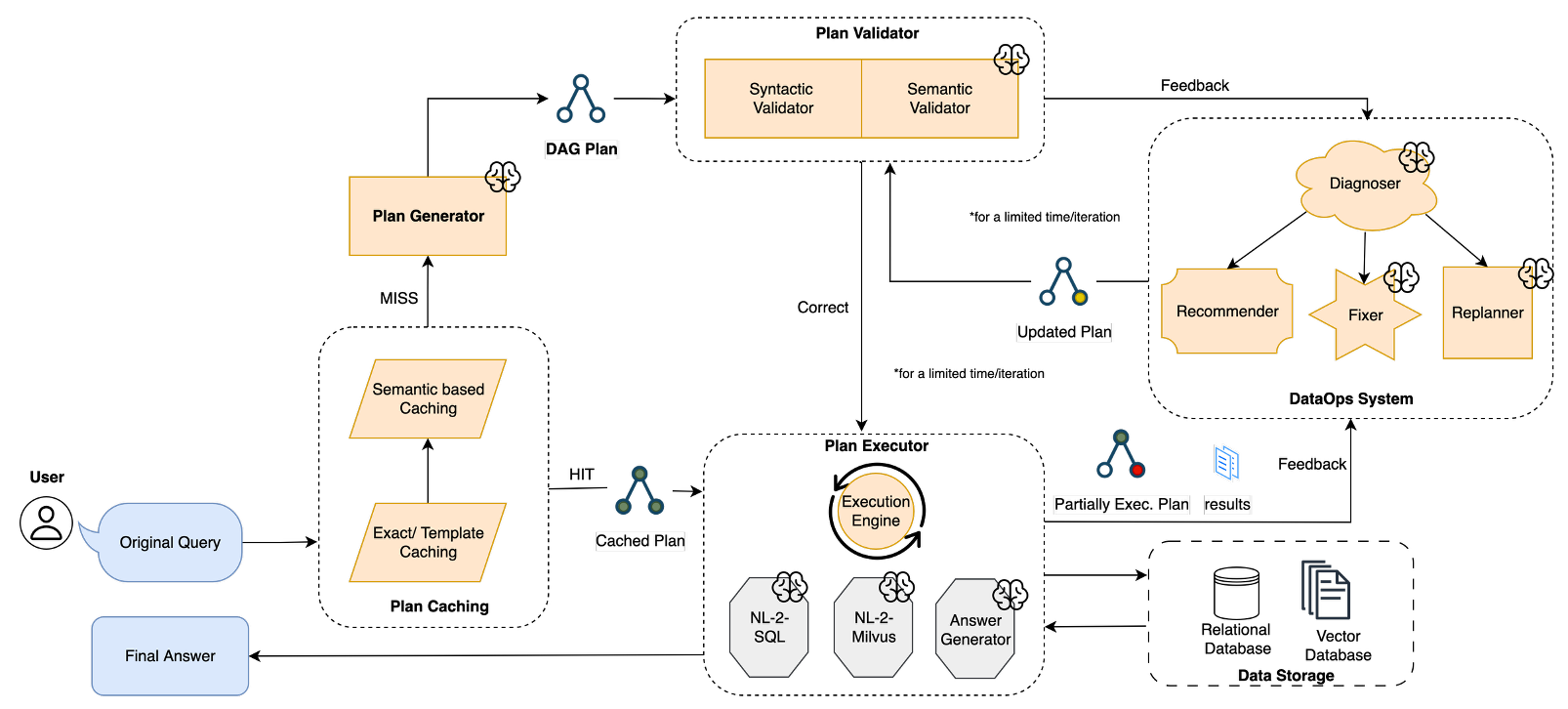}
    \caption{Agentic DAG-Orchestrated (A.DOT) Planner Architecture}
    \label{fig:main_framework}
\end{figure*}

\section{Hybrid Data Setting}
\label{sec:data}

We consider a hybrid data lake consisting of both structured and unstructured modalities: $\mathbf{D} = \{ D_S, D_U \}.$

\noindent\textbf{Structured Component.}
$D_S$ is modeled as a relational database comprising a set of tables:
\[
D_S = \{ T_1, T_2, \dots, T_m \},
\]
where each table $T_i$ has schema columns 
$\{ C_{i,1}, \dots, C_{i,k} \}$ 
and tuples 
$\{ R_{i,1}, \dots, R_{i,n} \}$. 
This component is managed in a relational engine.

\noindent\textbf{Unstructured Component.}
$D_U$ is a collection of documents:
\[
D_U = \{ d_1, d_2, \dots, d_p \}.
\]
Documents are chunked into semantically coherent units and embedded with both \emph{dense} and \emph{sparse} encoders. The resulting embeddings are stored in a vector index (e.g., Milvus):
\[
V_U = \{ v_1, v_2, \dots, v_q \},
\]
where each $v_\ell$ includes: (i) the chunk text, (ii) metadata (e.g., document id, timestamps), (iii) a dense vector for semantic similarity, and (iv) a sparse vector for lexical matching.

\noindent\textbf{Global Schema and Cross-links.}
We define a global schema $\mathcal{S}$ that captures both modalities:
\begin{itemize}
    \item \textbf{Relational schema:} column names, types, primary/foreign keys, constraints.
    \item \textbf{Vector-store schema:} collection fields and metadata keys (e.g., \texttt{document\_id}, \texttt{table\_name}, \texttt{row\_id}).
    \item \textbf{Cross-links:} mappings from vector metadata to relational keys, enabling traversal across modalities.
\end{itemize}

\noindent\textbf{Query Processing.}  
Given an NL query $Q$, the task is to retrieve only the relevant information from both $D_S$ and $D_U$, integrate it through multi-hop reasoning, and generate a coherent answer. Because such queries often span multiple modalities, execution may require switching repeatedly between structured and unstructured sources. Retrieval must be precise, with intermediate results selectively propagated across sources rather than relying on bulk extraction. Beyond correctness, the system must also preserve explicit data lineage at every stage, ensuring verifiability, compliance, and provenance tracking in enterprise settings.

\section{Agentic DAG-Orchestrated (A.DOT) Planner Framework}
\label{subsec:planner}


Figure~\ref{fig:main_framework} illustrates the architecture of the proposed Agentic DAG-Orchestrated (A.DOT) Planner. The framework comprises the following components:

\begin{enumerate}
    \item \textbf{Plan Caching:} Accelerates query processing by reusing DAG plans for similar NL queries.  
    \item \textbf{Plan Generator:} Decomposes the input query $Q$ into a DAG of atomic sub-queries.  
    \item \textbf{Plan Validator:} Applies syntactic and semantic checks to ensure executability and intent preservation.  
    \item \textbf{DataOps System:} Uses \emph{Diagnostic} and \emph{Remediation} agents to address invalid or incomplete plans through suggestions, quick fixes, or full replanning.

    \item \textbf{Plan Executor:} Coordinates execution of DAG nodes through an \emph{Execution Engine}, invoking sub-agents like NL-2-SQL, NL-2-Milvus, and answer generation, while preserving variable bindings and aggregating results. 
\end{enumerate}

Together, these modules enable robust, feedback-driven execution of queries across structured and unstructured data.



\subsection{Plan Caching}
\label{subsec:plan_caching}

To accelerate repeated or paraphrased queries, we maintain a repository that maps a \emph{query, context} key to validated plans:
\[
\mathcal{I} : \big(Q,\;\sigma(\mathcal{S}),\;\gamma\big) \;\mapsto\; \mathcal{P},
\]
where $\sigma(\mathcal{S})$ is a schema signature (e.g., hash of table/column metadata and cross-links) and $\gamma$ denotes execution context (e.g., user/role, policy flags). Each cached entry stores a parameterized DAG plan $\mathcal{P}$, and provenance summaries.

We support three retrieval strategies:
\begin{enumerate}
    \item \textbf{Exact caching:} returns $\mathcal{P}$ when $Q$ (after normalization) and the context $(\sigma(\mathcal{S}),\gamma)$ exactly match a cached key
    \item \textbf{Template caching:} stores parameterized templates with slots and a shared plan skeleton. Incoming $Q$ is matched via paraphrase-aware slot extraction; on a high-confidence match, the template is instantiated and the resulting plan is validated
    \item \textbf{Semantic caching:} indexes normalized queries with embeddings \cite{zhou2025paraphrase}. Given $Q$, we retrieve top candidates by semantic similarity and reuse a candidate plan iff structural validation succeeds
\end{enumerate}

The cache implements LRU eviction. On a hit, the system bypasses plan generation but still performs validation. Subquery-level caching (reusing DAG subgraphs across plans) is promising future work.

\subsection{Plan Generator}
\label{subsec:plan_generator}

The Plan Generator transforms an NL query $Q$ into a DAG plan $\mathcal{P} = (V,E)$ in a \emph{single} LLM pass. Each node $n \in V$ is an \emph{atomic} sub-query, answerable from exactly one data source. Nodes are annotated with:
\begin{itemize}
    \item \textbf{question:} NL sub-task description  
    \item \textbf{tool:} Target data source (\texttt{sql} for $D_S$, \texttt{vector} for $D_U$) 
    \item \textbf{label:} Symbolic variable storing $n$’s output 
    \item \textbf{should\_expose\_answer:} Boolean flag for early surfacing of intermediate results
\end{itemize}

Edges $(u \!\rightarrow\! v) \in E$ encode dependencies, where the output of $u$ is consumed by $v$. Independent nodes can be executed in parallel, while dependency edges coordinate information flow across structured and unstructured sources. An example DAG plan is shown in Appendix \ref{app:dag}.

\subsection{Plan Validator}
\label{subsec:plan_validator}

The validator ensures that a plan $\mathcal{P}$ is both executable under schema $\mathcal{S}$ and semantically aligned with query $Q$, intercepting errors before costly execution. It runs after cache hits or plan generation, returning feedback for diagnosis:
\[
V(\mathcal{P},\mathcal{S},Q) \rightarrow (\texttt{is\_valid}, \texttt{feedback}).
\]

\paragraph{Structural Checks.}
Verify schema compliance (all columns/tables are valid), variable hygiene (\texttt{\$var\_d.col} is not  dangling), node completeness (required fields present), and acyclicity of dependencies. Detailed rules and algorithm are provided in Appendix~\ref{app:val}.

\paragraph{Semantic Checks.}
Ensure intent preservation and execution feasibility: join keys respect types, and aggregates reference valid attributes. Open-source LLMs are used in an audit prompt to flag intent drift and suggest edits.

Dual validation (structural + semantic) captures both static inconsistencies and semantic errors, enabling safe, feedback-driven refinement.

\subsection{DataOps System}
\label{subsec:dataops}


When validation fails or execution raises exceptions, the \textit{DataOps System} is activated. It analyzes feedback and selectively triggers custom remediation:
\[
\texttt{DataOps}(\mathcal{P}, \mathcal{S}, H, F) \rightarrow \mathcal{A},
\]
where $\mathcal{P}$ is the current plan, $\mathcal{S}$ is the schema, $H$ is the edit history, $F$ is the feedback from validators or executors, and $\mathcal{A}$ denotes the resulting action. Depending on the failure type, the system may emit actionable recommendations for downstream decision-making, apply a targeted fix , or generate a revised plan. Formally, it consists of four submodules:


\begin{itemize}
    \item \textbf{Diagnoser:} Identifies root causes of failures (e.g., tool mismatch, unresolved variable, missing filters) by inspecting validator messages and execution traces.

    \item \textbf{Recommender:} Suggests non-executable recovery actions like server downtime, flagging such cases for user escalation or alternative execution paths.
    
    \item \textbf{Fixer:} Applies local, conservative edits (e.g., resolving renamed fields or fixing filter syntax) without altering DAG structure or prior node outputs.

    \item \textbf{Replanner:} Performs full or partial rewrites of the execution plan when deeper structural issues are detected. 
\end{itemize}

This modular enables progressive refinement: minor issues are resolved quickly by the Fixer, while more structural or semantic issues trigger the Replanner. By isolating diagnosis, recommendation, and remediation roles, the system ensures reliable recovery without full re-generation.

\subsection{Plan Executor}
\label{subsec:plan_executor}

The executor traverses the DAG in topological order, executing zero in-degree nodes in parallel while respecting dependencies. A variable store maintains bindings and exposes only minimal required attributes (e.g., \texttt{document\_id}s) to successor nodes to reduce payload overhead.

\paragraph{Key challenge: }
Intermediate query results, can contain large payloads. Naively passing these downstream can overwhelm memory or exceed context limits in generation models. To address this, the executor \emph{slims variables} by propagating only the necessary keys (e.g., IDs), which are then used to join, filter, or retrieve relevant subsets in dependent subqueries. This strategy ensures efficiency and correctness without compromising the logical flow of execution.

\paragraph{Source-specific execution:} 
\begin{itemize}
    \item \textbf{SQL nodes ($D_S$):} Translated via NL-to-SQL\cite{gao2023text}, with alias normalization and policy enforcement.
    \item \textbf{Vector nodes ($D_U$):} Handled via NL-to-Vector modules (e.g., Milvus), using sparse+dense retrieval and optional joins with SQL outputs.
\end{itemize}

\paragraph{Streaming and partial answers:} 
If a node has \texttt{should\_expose\_answer=true}, its result is immediately surfaced with \texttt{answer\_description}, while the rest of the DAG continues execution. Failures (e.g., SQL errors, vector timeouts) are logged and returned to the DataOps system as structured feedback.

\paragraph{Lineage and logging:} 
Each step records the executed operator, inputs, outputs, and data provenance (e.g., source rows or document spans), enabling downstream verification and auditability.

\paragraph{Final response generation:}
After all subqueries are executed, the system invokes a response synthesis module that composes the final answer using relevant subquery results. This step ensures coherence, eliminates redundancy, and produces a fluent, human-readable output.

\section{Experiments}
\label{sec:exp}

\subsection{Experimental Settings}
\label{subsec:exp_settings}

\subsubsection{Dataset:}
\label{subsec:datasets}

To evaluate the robustness and efficiency of A.DOT, we evaluate its performance on the HybridQA dataset \cite{hybridqa}, a challenging benchmark that demands multi-hop reasoning across both structured tables and unstructured text. Designed to emulate real-world enterprise scenarios, HybridQA tests a system’s ability to retrieve, link, and synthesize information from heterogeneous data sources.

The dataset is built from Wikipedia tables and their linked textual passages, requiring models to reason across both modalities. Each question often necessitates 1–2 hops between structured and unstructured content, traversing documents and table entries to derive the correct answer. For our evaluation, we focus on the development (dev) split, which consists of 3,466 question–answer pairs, carefully curated to reflect a wide range of reasoning challenges. We exclude the test set from our analysis, as the ground truth answers have not been publicly released.

\subsubsection{Implementation Details:}
\label{subsec:imp_details}

To adapt HybridQA for enterprise-grade benchmarking, we restructured its schema and storage formats to align with real-world data lake architectures. The original HybridQA dataset organizes structured content as JSON tables and links them to textual passages via cell-level annotations. However, this JSON-based format is impractical for large-scale enterprise systems, which typically rely on hybrid relational--vector backends. To bridge this gap, we recast HybridQA into a dual-store architecture:

\paragraph{Structured and Unstructured Storage:}

\begin{itemize}
    \item \textbf{Relational Tables:} All JSON tables are converted into a relational SQL backend (SQLDB). Each table preserves its original schema, maintaining column names, primary keys, and inter-row relationships.
    
    \item \textbf{Vector Store for Documents:} Unstructured textual passages, linked to table cells, are chunked into coherent segments and embedded into a Milvus vector database. This enables efficient dense and sparse retrieval.
\end{itemize}
\noindent
To ensure cross-modal connectivity:
\begin{itemize}
    \item We add a \texttt{document\_id} to rows in the SQL tables, establishing a connection to related document chunks.
    \item Each Milvus entry includes corresponding \texttt{document\_id}, allowing bidirectional traversal between table entries and their supporting evidence.
\end{itemize}

This hybrid design allows the system to reason across modalities, enabling NL queries to access and synthesize information from both relational and textual sources.




\paragraph{Execution Pipeline:}
The entire A.DOT pipeline is powered by the \textbf{LLaMA-3 70B} model and implemented using the LangGraph framework. Each component, as illustrated in Figure~\ref{fig:main_framework}, is modularized as an agent responsible for handling different stages of the process, from receiving the input query to producing the final answer.

\paragraph{Evaluation Metrics:} We evaluate A.DOT using \textbf{NL-centric metrics} from the Unitxt~\cite{bandel-etal-2024-unitxt} library:

\begin{enumerate}
    \item \textbf{Answer Correctness:} Measures whether the predicted answer is factually aligned with the reference. A binary label is produced via \textit{LLM-as-judge} using the Mistral Large 2 model, ensuring the response captures the correct meaning regardless of surface form.

    \item \textbf{Answer Completeness:} Assesses the extent to which the predicted answer covers all required information. Graded on a scale from \textit{Very Bad} to \textit{Excellent}, this metric highlights omissions and partial completions, again judged by Mistral Large 2.
\end{enumerate}

\subsection{Baselines}
\label{subsec:baselines}

To demonstrate the effectiveness of A.DOT, we compare it against established industry-standard agentic frameworks. Since most of these frameworks operate via tool calling (e.g., invoking external API calls or database queries), we reformulate our data sources into two core tools:
\begin{itemize}
    \item \textbf{SQL Retriever} Accepts an NL query, converts it into SQL, and retrieves results from the relational database
    \item \textbf{Milvus Retriever} Transforms NL query into a vector search and retrieves closest matching text chunks
\end{itemize}



Both tools are made available to all baselines, along with metadata describing the data lake, ensuring a fair comparison. The baselines include:

\paragraph{Standard RAG (Retrieval-Augmented Generation)} As a widely adopted baseline, the RAG setup retrieves both structured and unstructured evidence for each user query using the two retrievers. Retrieved results are concatenated and passed to the LLM in a single-pass generation step, without iterative reasoning or planning.

\paragraph{ReAct:}  We adopt the ReAct framework \cite{yao2023react} as a strong reasoning-first baseline. ReAct agents alternate between generating thoughts and invoking tools. At each step, the agent uses observations from tool outputs to guide subsequent reasoning. We provide ReAct with both SQL and Milvus Retrievers to align with our hybrid setup.

\paragraph{LLM Compiler} We also compare against the LLM Compiler framework \cite{kim2024llm}, which introduces a DAG-based planner for orchestrating tool calls. The planner can dynamically replan based on execution feedback and prior context, allowing more flexible coordination of SQL and Milvus retrievals. In our setup, both tools are provided to the planner, which produces a DAG-structured plan that can be executed in parallel or sequentially, depending on dependencies. This makes the LLM Compiler a stronger baseline, that more closely resemble A.DOT’s agentic execution.

\subsection{Evaluation Results}
\label{subsec:main_result}

\begin{table}[ht]
\centering
\renewcommand{\arraystretch}{1.1}
\setlength{\tabcolsep}{3pt}
\footnotesize
\begin{tabular}{@{}lcc@{}}
\toprule
\rowcolor{gray!20}
\textbf{Frameworks} & \textbf{Answer Correctness} & \textbf{Answer Completeness} \\
\midrule
LLM Compiler  & 27.8 & 30.8 \\
ReAct         & 40.2 & 44.3 \\
Standard RAG    & 56.2 & 62.3 \\
\textbf{A.DOT}        & \textbf{71.0} & \textbf{73.0} \\
\bottomrule
\end{tabular}
\caption{Comparison of various frameworks}
\label{tab:model_comparison}
\end{table}

Table~\ref{tab:model_comparison} presents a comparative evaluation of A.DOT against prominent agentic AI frameworks using the HybridQA benchmark. All models use \textbf{LLaMA-3-70B} for inference and \textbf{Mistral-Large} as the LLM-as-a-Judge (LLMaJ) for evaluation.

A.DOT significantly outperforms all baselines, improving upon the strongest baseline (Standard RAG) by \textbf{14.8\% in correctness} and \textbf{10.7\% in completeness}. This highlights its superior ability to perform multi-hop reasoning over hybrid structured and unstructured sources.

\paragraph{Standard RAG:}  
Performs relatively well when answers lie directly within unstructured documents. Since it uses the same retrieval backbone as A.DOT, it is able to fetch relevant chunks, but lacks deep reasoning capabilities and hence often fails on complex multi-hop queries spanning modalities.


\paragraph{ReAct:}
The ReAct agent leverages iterative reasoning and tool use but is capped at 20 rollout steps, a design choice aligned with our task requirements, which typically involve at most 2-hop reasoning. This safeguard ensures termination in cases of reasoning loops. While this avoids infinite loops, it limits recovery opportunities, leading to lower overall performance compared to RAG.

\paragraph{LLM Compiler:}  
While LLM Compiler introduces structured DAG-based planning and parallel execution to reduce latency, it underperforms due to over-reliance on relational tools. It frequently skips vector-based retrieval and lacks robust plan validation. Unlike A.DOT, which employs schema-aware validation and semantic feedback for robust replanning, LLM Compiler only responds to final node failures and is blind to mid-execution errors. It also lacks a modular diagnostic and remediation loop.

\paragraph{A.DOT (Ours):}  
A.DOT excels by integrating dynamic planning, fine-grained error handling, schema validation, and flexible evidence synthesis. It gracefully recovers from partial failures and ensures answer quality through streaming outputs, intermediate diagnostics, and answer validation.

\subsection{Ablation Study}
\label{ablation_studies}

\begin{table}[ht]
\centering
\renewcommand{\arraystretch}{1.3}
\begin{tabular}{@{}p{3.7cm}c c@{}}
\toprule
\rowcolor{gray!20}
\textbf{Model} & \makecell{\textbf{Answer} \\ \textbf{Correctness}} & \makecell{\textbf{Answer} \\ \textbf{Completeness}} \\
\midrule
Without Plan Validator \newline and DataOps System & 67.9 & 69.6 \\
Without DataOps System & 60.0 & 61.8 \\
Without Plan Validator & 68.0 & 69.6 \\
A.DOT & 71.8 & 74.3 \\
\bottomrule
\end{tabular}
\caption{Ablation study}
\label{tab:ablation_system}
\end{table}

Table~\ref{tab:ablation_system} presents results from the ablation study conducted on a 500 sample subset of the HybridQA development set. This analysis quantifies the individual and combined impact of two critical components in A.DOT: the Schema Validator and the DataOps System.

Interestingly, removing only the \textbf{DataOps System} results in the sharpest drop in performance (60.0\% Answer Correctness), as the system fails to recover from execution errors, even when these are correctly flagged by the Schema Validator. However, when both the \textbf{Schema Validator} and \textbf{DataOps System} are disabled, performance improves (67.9\%), which may seem counterintuitive. This is because, in the absence of the DataOps System, the Plan Validator halts execution upon detecting issues, but cannot repair the plan, resulting in premature termination. When both components are absent, the system proceeds unchecked and occasionally succeeds via direct execution, albeit without structural safeguards. These results highlight the tight coupling between validation and replanning: only when both operate together can the system detect, correct, and optimize plans effectively, leading to best performance.



We note that plan caching is excluded from this ablation, as its primary utility lies in avoiding redundant computation across recurring queries, common in enterprise use cases but rare in the diverse HybridQA benchmark.

\paragraph{Error Analysis:}  
We categorize observed failure cases into three broad types. Illustrative examples of these error categories are provided in Appendix~\ref{app:error}:
\begin{enumerate}
    \item \textbf{Intermediate Subquery Failures:} NL2SQL or NL2Milvus modules fail to retrieve necessary data.
    \item \textbf{Incorrect Data Source Assignment:} The planner misidentifies whether a subtask requires structured (SQL) or unstructured (Milvus) querying
    \item \textbf{Lack of Intrinsic Knowledge:} The system fails to incorporate commonsense knowledge  when reasoning across modalities
\end{enumerate}



\section{Deployment Plan}
\label{sec:deploy}
A.DOT is currently being evaluated for integration as the core retrieval module in \textit{IBM Watsonx.data Premium} \cite{think2025}, with the objective of enhancing answer accuracy and delivering verifiable outputs tailored to enterprise needs. This transition from research prototype to product is backed by planned user studies, extensive performance benchmarking, and validation over representative enterprise datasets. These efforts mark a clear trajectory toward real-world deployment, ensuring that A.DOT not only meets functional expectations but also aligns with the robustness and interpretability demands of production systems.

\section{Conclusion and Future Work}
\label{sec:conclusion}

In this paper, we introduced A.DOT, an enterprise-ready agentic framework for hybrid question answering over structured and unstructured data. Unlike traditional extractive QA systems, A.DOT synthesizes fluent, context-aware responses by integrating relevant data from relational databases and vector-based document stores.

Our architecture emphasizes modularity, fault-tolerant execution, and parallelism through a DAG-based planner. It incorporates intermediate result caching, robust schema validation and remediation mechanisms, and supports intermediate answer exposure during execution. Additionally, A.DOT maintains fine-grained lineage logs at each step, enhancing traceability, auditability, and interpretability, key requirements in enterprise-grade deployments. On the HybridQA benchmark, a challenging multi-hop QA dataset, A.DOT significantly outperforms agentic baselines such as ReAct, RAG, and LLM Compiler.



As a future research direction, we aim to extend A.DOT beyond single-turn queries to support multi-turn conversational settings, where maintaining context history and resolving user intent are critical. The system’s modular design further allows seamless integration of additional modalities, such as graph databases, image-based knowledge, thereby broadening its applicability across a wide range of real-world hybrid data environments.


\bibliography{aaai2026}

\appendix

\section{Related Works}
\label{app:rel_works}

Recent work has pushed multi-hop Hybrid Question Answering (HybridQA) – where queries require combining structured data (tables) with unstructured text – using advanced LLMs and novel pipelines. Traditional systems often relied on supervised retriever-reader models[\cite{sun2018open},\cite{muller-etal-2021-tapas}], but these demand large training data. S3HQA\cite{lei-etal-2023-s3hqa} fine-tuned a retriever, a hybrid selector, and a generator to reach state-of-the-art accuracy on the HybridQA benchmark\cite{hybridqa}. Recent approaches, leverage in-context learning and multi-step reasoning[\cite{guan2024mfort},\cite{yu2025tablerag},\cite{feng-etal-2022-multi}] to reduce training needs. One such method is TTQA-RS\cite{bardhan2024ttqa}, a prompt-based pipeline that explicitly decomposes questions and retrieves evidence iteratively. Another zero-shot method is the Hybrid Graph approach\cite{agarwal2025hybrid}. Instead of serial prompting, it constructs a unified graph of all candidate table cells and text snippets and prunes it based on the question. Relative to A.DOT, which plans explicit query operations, the Hybrid Graph is more of a dynamic context builder; it excels at filtering relevant facts but does not yield a reusable “plan” beyond the immediate context – the final reasoning still happens implicitly in the LLM’s forward pass.

In parallel, the NLP community has explored using LLMs themselves as planning and acting agents to tackle multi-step information tasks. \cite{yao2023react} propose ReAct, which interleaves the LLM’s chain-of-thought reasoning with actions such as external queries. This allows the model to fetch information mid-reasoning and update its plan, reducing hallucinations in multi-hop QA. Other work treats the LLM as a high-level controller. HuggingGPT\cite{shen2023hugginggpt} uses ChatGPT to decompose a complex query into sub-tasks, invoke appropriate expert models (e.g. vision or text QA models from HuggingFace) for each, then integrate their outputs to produce the answer. A complementary line of research makes reasoning explicit through planning. BLENDSQL\cite{glenn2024blendsql}  compiles hybrid QA into SQL-like queries that deterministically join structured and unstructured sources, yielding interpretable, reusable plans and lower context overhead. Similarly, LLMCompiler\cite{kim2024llm} generalizes this idea into parallel DAG execution of function calls, improving efficiency and throughput. Yet, while LLMCompiler optimizes when to execute, it leaves the what under-specified. By contrast, A.DOT emphasizes schema-aware validation, caching, and reusability—bridging the gap between interpretable query planning (as in BLENDSQL) and efficient orchestration (as in LLMCompiler).

Frameworks like LangChain\cite{langchain} further popularize such agent-based orchestration, letting an LLM choose tool APIs (search, calculators, DB queries, etc.) in an interactive loop. In summary, our approach integrates the strengths of prior HybridQA systems (tight integration of heterogeneous data) with the adaptability of LLM-based agents, using a DAG-based planner to achieve more reliable and efficient multi-hop reasoning across hybrid data lakes.

\begin{figure}[h]
  \centering
  \includegraphics[width=0.46\textwidth,height=0.8\textheight,keepaspectratio]{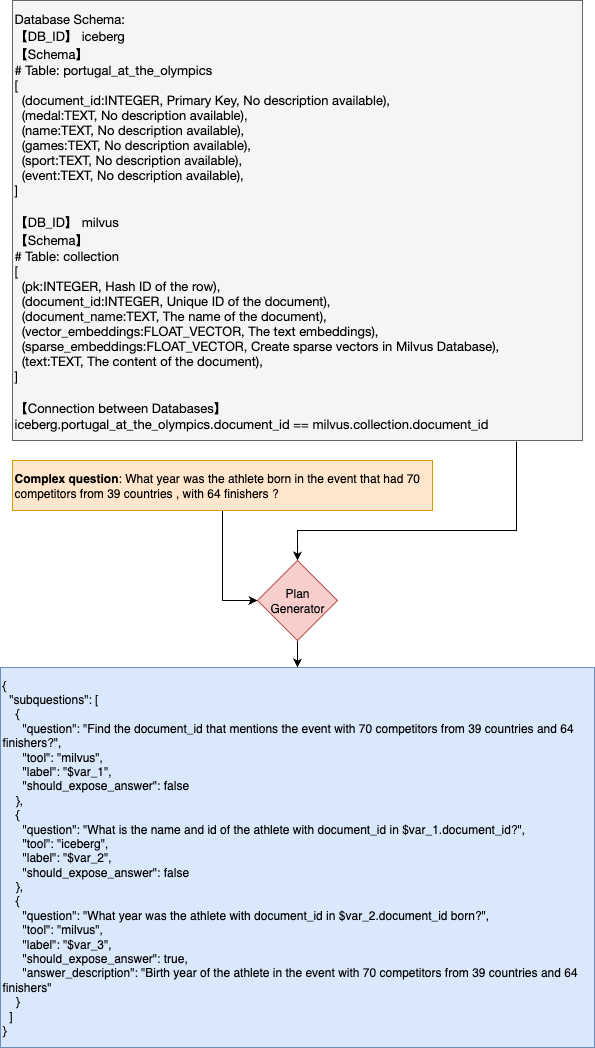}
  \caption{Example DAG plan}
  \label{fig:dag_plan}
\end{figure}

\section{DAG Example}
\label{app:dag}
Consider the question, \emph{What year was the athlete born in the event that had 70 competitors from 39 countries, with 64 finishers?}. Figure ~\ref{fig:dag_plan}, shows the corresponding DAG plan. The Plan Generator produces three atomic sub-queries:  
(1) retrieve the relevant event ID from the document collection ($D_U$),  
(2) use this ID to fetch athlete details from the structured table ($D_S$), and  
(3) resolve the athlete’s birth year back in $D_U$.  
The final node is annotated with \texttt{should\_expose\_answer=true}, enabling the system to surface the athlete’s birth year.

\section{A.DOT Plan Validation}
\label{app:val}

\begin{algorithm}
\caption{Execution Plan Validation}
\begin{algorithmic}[1]

\STATE \textbf{Input:} $S$ (Database schema), $P$ (Execution plan with list $P.\mathsf{subquestions}$)

\STATE {\ }

\STATE \textbf{/* 1. Plan Structural Specification */}
\IF{${subquestions} \notin P$ \textbf{or} $P.subquestions$ not a list}
    \STATE \textbf{Error}
\ENDIF

\FOR{each $q_i$ in $P.\mathsf{subquestions}$, $i = 1$ to $n$}
    \IF{$q_i.\mathsf{status} = \mathsf{executed}$}
        \STATE \textbf{Continue}
    \ENDIF
    \STATE Require $\mathsf{question}$, $\mathsf{tool}$, $\mathsf{label}$, and $\mathsf{should\_expose\_answer}$ exist
    \STATE Assert $\mathsf{question}$ is a non-empty string
    \STATE Assert $\mathsf{tool} \in \{\mathsf{iceberg}, \mathsf{milvus}\}$
    \STATE Assert $\mathsf{label}$ matches \texttt{\$var\_int}
    \STATE Assert $\mathsf{should\_expose\_answer} \in \{\mathsf{True}, \mathsf{False}\}$
    \IF{$\mathsf{should\_expose\_answer} = \mathsf{True}$}
        \STATE Require $\mathsf{answer\_description} \neq \emptyset$
    \ENDIF
\ENDFOR

\STATE Require at least one $q_i$ with $\mathsf{should\_expose\_answer} = \mathsf{True}$

\STATE {\ }

\STATE \textbf{/* 2. Dependency-Expression Hygiene */}
\STATE Pattern gets \texttt{\$var\_d[.c]}

\FOR{each non-executed $q_i$}
    \FOR{each match $(v_d, c)$ in question}
        \STATE $d \gets$ integer after \texttt{\$var\_}
        \IF{$d > n$}
            \STATE \textbf{Error: Unknown variable}
        \ENDIF
        \IF{$c \neq \emptyset$ \textbf{and} $c \notin S$ \textbf{and} $c \notin$ partialResult[\texttt{\$var\_d}]}
            \STATE \textbf{Error: Unknown column}
        \ENDIF
    \ENDFOR
\ENDFOR

\STATE {\ }

\STATE \textbf{/* 3. Acyclic Intralogic Dependencies */}
\STATE Build directed graph $G$
\STATE Nodes are labels of $q_i$ (strip leading \texttt{\$})
\STATE Add edge $(u \rightarrow v)$ if $q_u$ references variable \texttt{\$var\_v}
\IF{DFS detects cycle in $G$}
    \STATE \textbf{Error: Cyclic dependency}
\ENDIF

\STATE {\ }

\STATE \textbf{/* 4. Result */}
\STATE $\mathsf{VALID} \gets \mathsf{True}$ if no errors recorded
\STATE Output: $\mathsf{VALID}$, feedback: $\mathsf{list\_of\_errors}$

\end{algorithmic}
\end{algorithm}

Given an execution plan $P$ with subquestions and a database schema $S$, our validation algorithm proceeds in four phases:

\textbf{(1) Structural Specification.} Checks that $P.\mathsf{subquestions}$ is a list, and each subquestion $q_i$ contains the required fields (\texttt{question}, \texttt{tool}, \texttt{label}, \texttt{should\_expose\_answer}). It ensures \texttt{tool} belongs to a known backend (\texttt{iceberg}, \texttt{milvus}) and \texttt{label} matches a valid variable pattern. If \texttt{should\_expose\_answer} is \texttt{True}, the corresponding \texttt{answer\_description} must be present. At least one subquestion must have \texttt{should\_expose\_answer = True}.

\textbf{(2) Dependency-Expression Hygiene.} Ensures all variable references of the form \texttt{\$var\_d[.c]} point to valid subquestions and columns. It raises an error if a referenced subquestion index $d$ is out of bounds or the column $c$ is undefined in $S$ or the partial result of $q_d$.

\textbf{(3) Acyclic Dependencies.} Builds a directed graph of variable dependencies across subquestions. A depth-first search is used to detect and reject cyclic dependencies.

\textbf{(4) Result.} The plan is marked \texttt{VALID} if no errors occur. Otherwise, a list of errors is returned.

Each subquestion is treated atomically and expected to execute on a single data source, avoiding any multi-source orchestration.

\section{Error Scenarios}
\label{app:error}
The Tables \ref{tab:example1}, \ref{tab:example2} and \ref{tab:example3} cover the error cases that were observed in the proposed A.DOT framework.

\begin{table*}[h]
\centering
\begin{tabular}{|p{3cm}|p{11cm}|}
\hline
\textbf{question\_id} & 03f0b21f92ee77b2 \\
\hline
\textbf{question} & What is the state represented by the teen whose home town is one of the gateways to the Great Smoky Mountains National Park? \\
\hline
\textbf{table\_id} & Miss\_Teen\_USA\_special\_awards\_1 \\
\hline
\textbf{answer-text} & Tennessee \\
\hline
\textbf{Plan Generator} & 
\begin{minipage}[t]{\linewidth}
\begin{lstlisting}[basicstyle=\small\ttfamily, breaklines=true]
{
"subquestions": [
  {
    "question": "Find the document_id of the teen whose hometown is one of the gateways to the Great Smoky Mountains National Park?",
    "tool": "milvus",
    "label": "$var_1",
    "should_expose_answer": false
  },
  {
    "question": "What is the state represented by the teen with document_id in $var_1.document_id?",
    "tool": "iceberg",
    "label": "$var_2",
    "should_expose_answer": true,
    "answer_description": "State represented by the teen whose hometown is one of the gateways to the Great Smoky Mountains National Park"
  }
]
}
\end{lstlisting}
\end{minipage} \\
\hline
\textbf{Milvus response} & 
\begin{minipage}[t]{\linewidth}
\begin{lstlisting}[basicstyle=\small\ttfamily, breaklines=true]
{
"reasoning": "The query asks for the document_id of the teen whose hometown is one of the gateways to the Great Smoky Mountains National Park. After analyzing the chunks, I found that none of the provided chunks explicitly mention the Great Smoky Mountains National Park or its gateways. However, I can infer that the gateways to the Great Smoky Mountains National Park include cities like Gatlinburg, Pigeon Forge, and Sevierville in Tennessee, as well as Cherokee, North Carolina. Since none of the chunks mention these cities or the park, I couldn't find a direct match. The response is based on the analysis of the provided chunks.",
"document_id": []
}
\end{lstlisting}
\end{minipage} \\
\hline
\end{tabular}
\caption{In this example, the Great Smoky Mountains National Park is not available from the milvus chunks so the vectorstore retriever fails to gather the information or the document\_id to pass it to the next iceberg query.}
\label{tab:example1}
\end{table*}

\vspace{1cm}

\begin{table*}[h]
\centering
\begin{tabular}{|p{3cm}|p{11cm}|}
\hline
\textbf{question\_id} & 15dd756a1e80bd1e \\
\hline
\textbf{question} & Where is the venue of the club that won the Bathurst 12 Hour located? \\
\hline
\textbf{table\_id} & Sport\_in\_Queensland\_1 \\
\hline
\textbf{Table Schema} & 
\begin{minipage}[t]{\linewidth}
\begin{lstlisting}[basicstyle=\small\ttfamily, breaklines=true]
# Table: sport_in_queensland
[
  (document_id:INTEGER, Primary Key, No description available),
  (club:TEXT, No description available),
  (league:TEXT, No description available),
  (venue:TEXT, No description available),
  (established:TEXT, No description available),
  (premierships:TEXT, No description available),
]
\end{lstlisting}
\end{minipage} \\
\hline
\textbf{answer-text} & Willowbank \\
\hline
\textbf{Plan Generated} & 
\begin{minipage}[t]{\linewidth}
\begin{lstlisting}[basicstyle=\small\ttfamily, breaklines=true]
{
"subquestions": [
  {
    "question": "Find the document_id of the club that won the Bathurst 12 Hour?",
    "tool": "milvus",
    "label": "$var_1",
    "should_expose_answer": false
  },
  {
    "question": "What is the venue of the club with document_id in $var_1.document_id?",
    "tool": "iceberg",
    "label": "$var_2",
    "should_expose_answer": true,
    "answer_description": "Venue of the club that won the Bathurst 12 Hour"
  }
]
}
\end{lstlisting}
\end{minipage} \\
\hline
\textbf{Predicted Answer} & The venue of the club that won the Bathurst 12 Hour is located at Queensland Raceway. \\
\hline
\end{tabular}
\caption{In this example, the venue information is available in the structured data, but still the answer-text retrieves the same information from the unstructured context.}
\label{tab:example2}
\end{table*}

\vspace{1cm}

\begin{table*}[h]
\centering
\begin{tabular}{|p{3cm}|p{11cm}|}
\hline
\textbf{question\_id} & 05625b361c2d12e9 \\
\hline
\textbf{question} & How many games were developed by a company that would become Vivendi games in 2006? \\
\hline
\textbf{table\_id} & Index\_of\_MS-DOS\_games\_(Q)\_0 \\
\hline
\textbf{answer-text} & 4 \\
\hline
\textbf{Plan Generated} & 
\begin{minipage}[t]{\linewidth}
\begin{lstlisting}[basicstyle=\small\ttfamily, breaklines=true]
{
"subquestions": [
  {
    "question": "Find the document_id of the club that won the Bathurst 12 Hour?",
    "tool": "milvus",
    "label": "$var_1",
    "should_expose_answer": false
  },
  {
    "question": "What is the venue of the club with document_id in $var_1.document_id?",
    "tool": "iceberg",
    "label": "$var_2",
    "should_expose_answer": true,
    "answer_description": "Venue of the club that won the Bathurst 12 Hour"
  }
]
}
\end{lstlisting}
\end{minipage} \\
\hline
\textbf{Predicted Answer} & The company that would become Vivendi Games in 2006 developed 3 games. \\
\hline
\end{tabular}
\caption{In this example, the milvus retriever only retrieves the 3 chunks instead of 4}
\label{tab:example3}
\end{table*}
\end{document}